%% file: main.tex
\documentclass[10pt,twocolumn,letterpaper]{article}

\usepackage[pagenumbers]{cvpr}

\usepackage{multirow} 
\usepackage{booktabs} 

\usepackage{xr-hyper}     
\input{preamble}

\definecolor{cvprblue}{rgb}{0.21,0.49,0.74}
\usepackage[pagebackref,breaklinks,colorlinks,allcolors=cvprblue]{hyperref}
\usepackage{comment}

\title{Complet4R: Geometric Complete 4D Reconstruction}

\author{
Weibang Wang$^{1, *}$ \quad
Kenan Li$^{1, *}$ \quad
Zhuoguang Chen$^{1, 2, *}$ \quad
Yijun Yuan$^{1, \dagger}$ \quad
Hang Zhao$^{1, 2, 3, \dagger}$
\\[0.5em]
$^1$IIIS, Tsinghua University \ \ \
$^2$Shanghai Artificial Intelligence Laboratory \ \ \ 
$^3$Shanghai Qi Zhi Institute 
\vspace{-1em}
}

\newcommand{\tok}{\mathbf{t}}
\newcommand{\bg}{\mathbf{g}}

\newcommand{\ourwork}{Complet4R\xspace}
\newcommand{\ourtask}{Geometric Complete 4D Reconstruction\xspace}

\begin{document}

\twocolumn[{
\renewcommand\twocolumn[1][]{#1}
\maketitle

\begin{center}
\vspace{-3mm}
    \includegraphics[width=\linewidth]{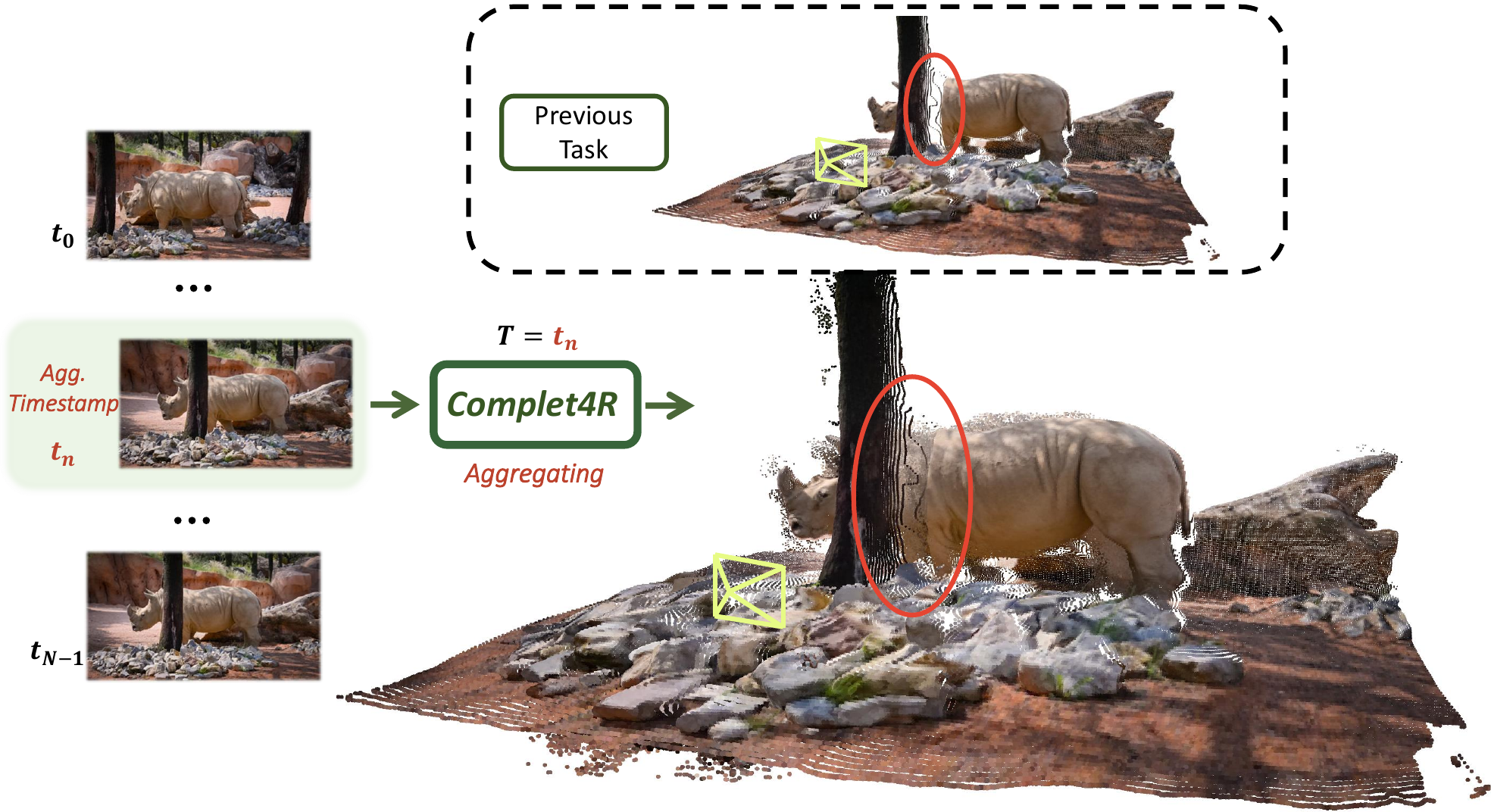}
    \captionof{figure}{\textbf{Complete and Consistent 4D Reconstruction.} Our model, \ourwork, aggregates 3D point maps from all frames into a specific timestamp, forming a complete geometric representation that recovers occluded regions visible from other views. By alternating the aggregated timestamp, our method achieves complete and consistent 4D reconstruction, producing temporally coherent and geometrically complete representations directly from sequential video input.}
    \label{fig:teaser}
\end{center}
}]

{\let\thefootnote\relax\footnotetext{{$^*$ Equal contribution.} $^\dagger$Equal advising.}}

\input{sec/0_abstract}    
\input{sec/1_intro}

\input{sec/2_rela}
\input{sec/3_method}
\input{sec/4_exp}
\input{sec/5_conclusion}

\input{sec/6_acknowledgment}
{
    \small
    \bibliographystyle{ieeenat_fullname}
    \bibliography{main}
}
 
\input{sec/X_suppl}

\end{document}

%% file: preamble.tex
\makeatletter
\renewcommand{\paragraph}{%
    \@startsection{paragraph}{4}%
    {\z@}{-0.5em}{-0.5em}%
    {\normalfont\normalsize\bfseries}%
}
\makeatother

\newcommand*{\V}{\mathbf}

%% file: sec/0_abstract.tex
\begin{abstract}

We introduce \ourwork, a novel end-to-end framework for \ourtask, which aims to recover temporally coherent and geometrically complete reconstruction for dynamic scenes. Our method formalizes the task of Geometric Complete 4D Reconstruction as a unified framework of reconstruction and completion, by directly accumulating full contexts onto each frame. Unlike previous approaches that rely on pairwise reconstruction or local motion estimation, \ourwork utilizes a decoder-only transformer to operate all context globally directly from sequential video input, reconstructing a complete geometry for every single timestamp, including occluded regions visible in other frames. Our method demonstrates the state-of-the-art performance on our proposed benchmark for \ourtask and the 3D Point Tracking task. Code will be released to support future research.

\end{abstract}

%% file: sec/1_intro.tex
\section{Introduction}
\label{sec:intro}

Modeling the 3D world has long been a central problem~\cite{marr2010vision} in both computer vision and robotics.
Classical structure-from-motion (SfM) and simultaneous localization and mapping (SLAM) approaches reconstruct static environments by combining point matching, triangulation, and bundle adjustment~\cite{seitz2006comparison, agarwal2011building, mur2015orb}.
While highly effective for rigid scenes, these formulations break down in the presence of dynamics: moving objects violate the rigidity assumption, often appearing distorted or being entirely removed from the reconstruction. Traditional methods tend to treat dynamic regions as noise to maintain spatial consistency, which inevitably degrades reconstruction quality when motion is prevalent.
However, dynamics are far more than noise. They encode rich spatial and temporal correlations that are vital for understanding the real world. By accurately modeling the dynamics, 3D perception can be extended into the spatiotemporal (4D) domain, where consistency must hold not only across space but also over time. Such 4D representations provide a foundation for higher-level reasoning about the physical world, enabling the development of world models and deeper insights into causality.

Early efforts to handle scene dynamics were largely built on 2D cues, such as optical flow~\cite{black1993framework} and long-range pixel correspondences~\cite{rubinstein2012towards}. Although these techniques yield dense trajectories in the image plane, they cannot resolve the fundamental ambiguities of projection: a single 3D point may map to different pixels under changing viewpoints, while distinct 3D points may collapse into one pixel through occlusion or perspective. Consequently, recovering accurate geometry or enforcing temporal consistency across views remains formidable.

Recent advances in 3D tracking and dynamic reconstruction have lifted motion analysis from the image plane to 3D space.
By estimating dense point maps and establishing frame-to-frame correspondences~\cite{wang2024dust3r, leroy2024grounding}, these methods represent dynamic scenes as temporally coherent 3D videos~\cite{zhang2024monst3r}. To further tame non-rigid deformation, some approaches decompose the task into a set of rigid-motion sub-problems~\cite{feng2025st4rtrack}, gaining local stability and interpretability without sacrificing global consistency.

Despite the progress, the dominant paradigm remains pairwise: each step reasons over only two frames.
This shortsighted matching makes errors accumulate through the sequence and offers no lever for global spatiotemporal regularization.
Moreover, their intermediate representations, point maps or 3D flow fields, are still frame-centric, limiting their capacity to represent the continuous evolution of real-world dynamics. 
Consequently, they can hardly distill the full sequence into a single, globally consistent 4D structure that fuses shape and motion across time directly.

Unlike prior approaches that either ignore temporal consistency~\cite{zhang2024monst3r} or only perform frame-level tracking~\cite{feng2025st4rtrack}, our framework offers a new concept to explore spatiotemporal coherence by completing 4D across the whole stream.
Which is, for every single frame, the 3D geometry is completed from all observations, so spatiotemporal coherence is baked in from the outset rather than retrofitted.

Our method provides a foundational representation for reasoning about real-world dynamics and serves as a step toward constructing physically grounded models.
Our main contributions are:
\begin{enumerate}
\item We introduce a new problem, complete consistent 4D scene reconstruction, which aims to recover temporally coherent and geometrically complete 4D representations from dynamic scenes.
\item To tackle this challenging problem, we propose \ourwork, a novel framework that effectively integrates motion information into a consistent, complete 4D geometric representation.
\item Extensive experiments demonstrate \ourwork achieves state-of-the-art or competitive performance on both the proposed \ourtask and its sibling 3D Point Tracking task benchmarks.
\end{enumerate}

%% file: sec/2_rela.tex
\section{Related Work}
\label{sec:related_work}

\subsection{Camera Estimation and Scene Reconstruction}
Joint prediction of camera poses and scene geometry has long been a classical problem in computer vision. Traditional Structure-from-Motion (SfM) methods like~\cite{schonberger2016structure}, VGGSfM~\cite{wang2024vggsfm} and Simultaneous Localization and Mapping (SLAM) methods like MonoSLAM~\cite{davison2007monoslam}, ORB-SLAM~\cite{mur2015orb} require extensive parameter tuning and incur high computational costs, which limit practical deployment.

Recently, a variety of learning-based approaches for monocular and video depth estimation have brought new opportunities. For example, DUSt3R~\cite{wang2024dust3r} predicts camera poses and scene geometry from image pairs, but its computationally intensive post-processing limits scalability to large scenes. Building on this trend, several notable works, such as VGGT~\cite{wang2025vggt}, enable high-precision prediction of camera poses, depth maps, and 2D tracking through a single forward pass of multi-view inputs, representing a new paradigm of data-driven 3D scene understanding.

However, these methods target static scenes, while dynamic scene reconstruction remains challenging. Existing approaches, such as R-CVD~\cite{kopf2021robust}, CasualSAM~\cite{zhang2022structure}, and MegaSAM~\cite{li2025megasam}, jointly optimize camera parameters and dense depth maps with depth priors. A notable follow-up to DUSt3R~\cite{wang2024dust3r}, St4RTrack~\cite{feng2025st4rtrack} uses a dual-branch architecture for dynamic reconstruction and tracking, but its pair-wise input still limits long-range predictions.

Therefore, to introduce a globally consistent model that addresses above limitations is of great interest to the field. 

\subsection{4D Reconstruction}

The 4D reconstruction problem has been extensively studied in non-rigid reconstruction. Early approaches relied on RGB-D sensors~\cite{zollhofer2014real, newcombe2015dynamicfusion, innmann2016volumedeform, dou2016fusion4d, bozic2020deepdeform} or strong hand-crafted priors~\cite{bregler2000recovering, kumar2017monocular, ranftl2016dense, russell2014video, fragkiadaki2014grouping}, while later works leveraged monocular depth priors~\cite{zhang2022structure, luo2020consistent} for better generalization. Recent advances in neural rendering~\cite{mildenhall2021nerf} and Gaussian Splatting~\cite{kerbl20233d} have further improved dynamic scene reconstruction, yet most methods~\cite{li2022neural, song2023nerfplayer, yang2025widerange4d} focus on novel view synthesis and photorealistic rendering rather than recovering physically meaningful geometry, thus neglecting geometric 4D consistency, an increasingly critical property for spatial visual intelligence. Many approaches also depend on auxiliary priors or assumptions~\cite{wang2024shape}, including known camera parameters, motion cues~\cite{dreamscene4d}, rigidity constraints, and even semantics from large language models~\cite{xu2024comp4d,   bahmani2024tc4d, zeng2024trans4d}. More recent efforts jointly infer camera poses, persistent geometry, and 3D trajectories from monocular videos~\cite{lei2025mosca, qingming2025modgs, wang2025continuous}, but they still require per-scene optimization and off-the-shelf priors. Although ~\cite{feng2025st4rtrack, han2025d, sucar2025dynamic} reveal the potential of complete dynamic reconstruction from multiple observations, their pairwise correspondence backbone limits performance to pairwise tracking and frame-wise reconstruction. 

In contrast, we propose a feed-forward framework that achieves 4D complete reconstruction. It reconstructs the scene and objects not only from the current frame, but also completes them from all observed parts in monocular videos, enabling geometrically consistent 4D representations and moving toward a complete physically grounded understanding of dynamic scenes.

\subsection{3D Point Tracking}

The tracking task was first introduced in Particle Video~\cite{sand2008particle}. Early optical flow and scene flow methods~\cite{horn1981determining,hur2020self}, which typically estimate motion only between adjacent frames, struggle to handle long video sequences. CoTracker~\cite{karaev2024cotracker} was the first to leverage a transformer architecture with joint attention to enable tracking through occlusions. Following this, BootsTAPIR~\cite{doersch2024bootstap} and CoTracker3~\cite{karaev2024cotracker3} explored the use of unlabeled data to further enhance performance. However, many existing long-sequence tracking methods~\cite{karaev2024cotracker, karaev2024cotracker3, xiao2025spatialtrackerv2} operate only in the camera coordinate system, making it difficult to achieve globally consistent tracks.

Recently, increasing attention has been given to 3D tracking. For example, SpatialTrackerV2~\cite{xiao2025spatialtrackerv2} achieves pixel-level tracking by jointly estimating camera poses and monocular depth. Another notable method is MVTracker~\cite{xu2025mitracker}, which extracts multi-level scene features to achieve high-precision 3D tracking across multiple views. However, its reliance on pre-calibrated camera parameters and ground-truth depth maps significantly restricts its applicability in real-world scenarios. While our model can also support the 3D tracking task without those limitations.

%% file: sec/3_method.tex
\section{\ourtask}

In this section, we introduce a novel task named \textbf{\ourtask}, which aims to reconstruct the complete geometry, including regions that are occluded in the current frame but visible in other frames. In essence, the objective is to \textit{aggregate information across time} so that all observed points, from both dynamic objects and static backgrounds, contribute to the reconstruction of a specific target timestamp. The key challenge of this task lies in accurately estimating the motion of dynamic points between timestamps, even when they are occluded in the target frame. To address this challenge, we propose \textbf{\ourwork}, a decoder-only transformer framework that globally reasons over sequential video inputs to achieve geometrically consistent and complete 4D reconstruction.

In the following subsections, we first define the task formulation and clarify its differences from prior related tasks. We then describe the architecture of \ourwork and its training strategy in detail.
\begin{figure}[t!]
    \centering
    \includegraphics[width=0.48\textwidth]{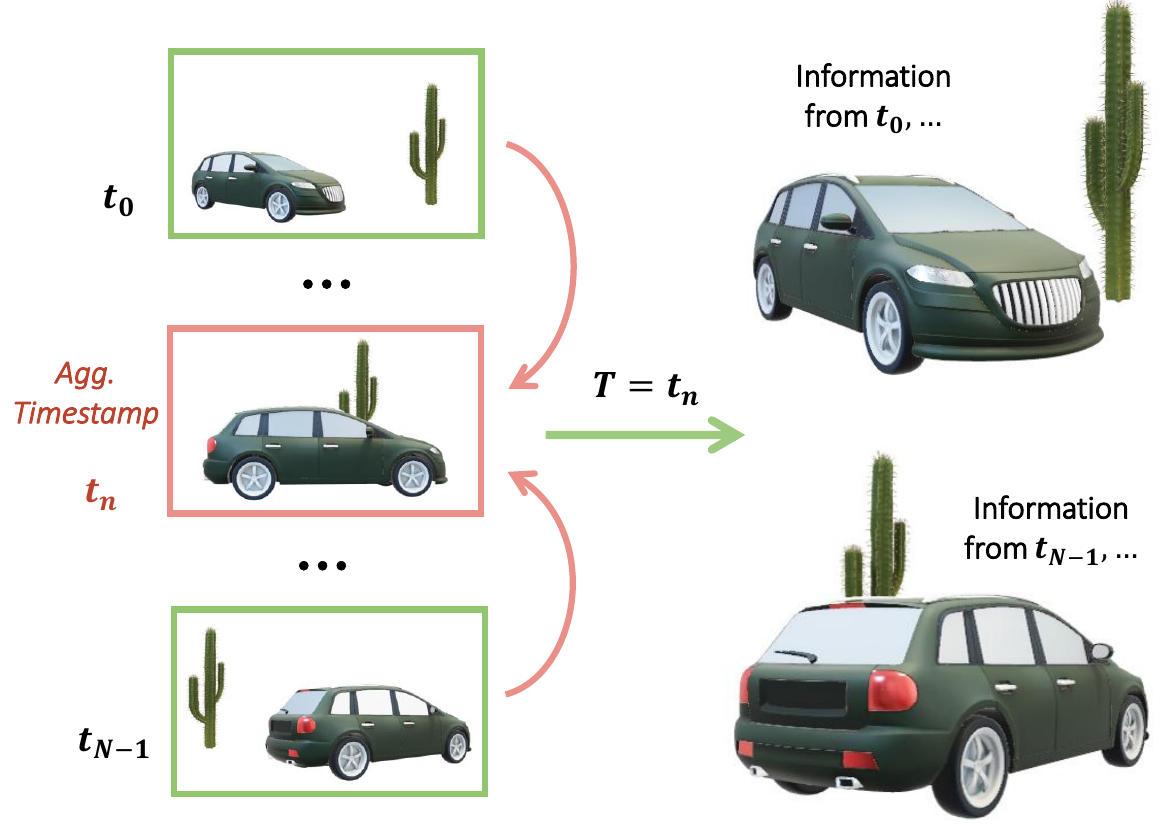}
    \caption{Geometric 4D complete reconstruction from observations. 
    Given input frames, Complet4R aggregates contextual information across all timestamps. Consequently, at each timestamp $T$, the reconstructed scene incorporates the geometry from frame $T$ along with complementary information from all other frames.
    }
    \label{fig:concept}
\end{figure}

\begin{figure*}[t!]
    \centering
    \includegraphics[width=1.0\textwidth]{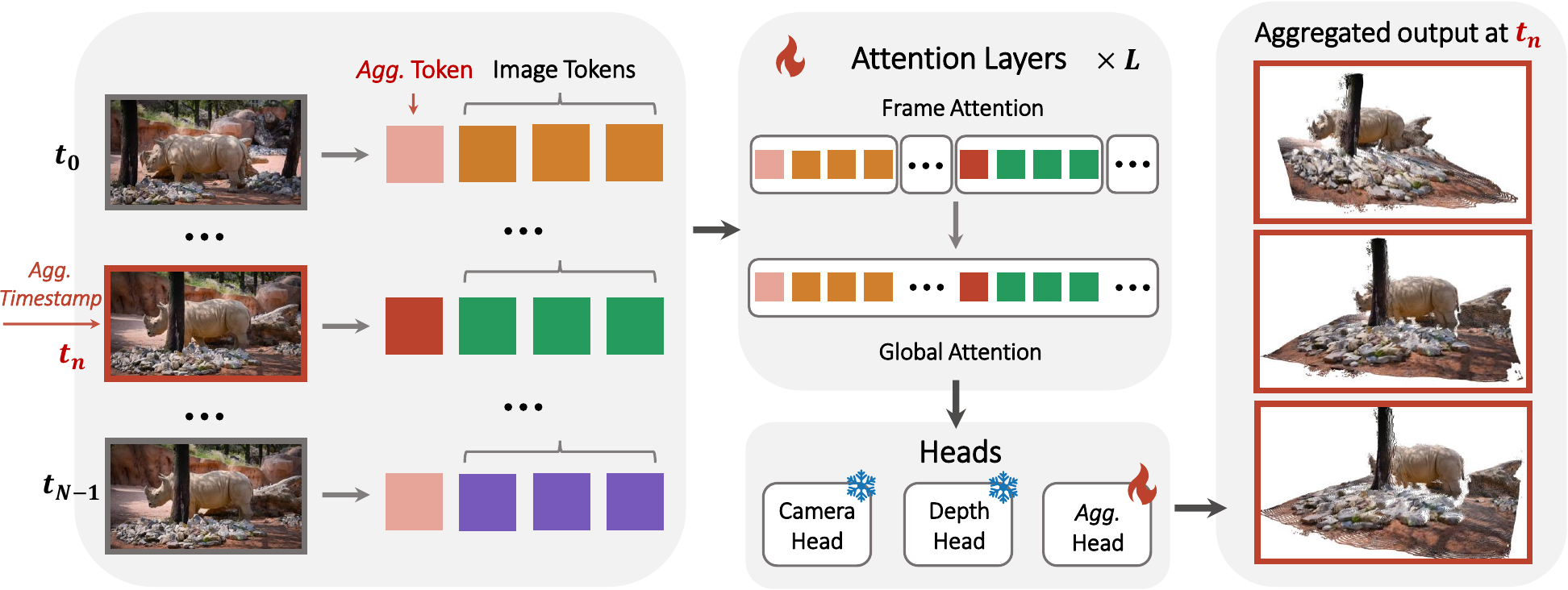}
    \caption{\textbf{Architecture Overview.} By concatenating special aggregation tokens, \ourwork identifies the specific timestamp for aggregation. The Aggregation head then outputs the positions of 3D points from other views at this timestamp, aggregating 3D point maps across frames to form a complete geometric representation.}
    \label{fig:main}
\end{figure*}
\subsection{Task Formulation}
\paragraph{Revisiting Prior Task Definitions.}

Prior 4D reconstruction works, such as MonST3R~\cite{zhang2024monst3r}, have made notable progress in modeling dynamic scenes from monocular videos. These methods typically regress the \textit{visible} pointmap corresponding to each timestamp, focusing on reconstructing the geometry observable in the current frame. However, they overlook that different portions of dynamic objects or static regions may only become visible at other timestamps due to occlusions or motion. Consequently, these approaches provide \textit{spatiotemporally incomplete} geometry representations.

\paragraph{Our Definition: Geometric Complete 4D Reconstruction.}
We propose a novel task named Geometric Complete 4D Reconstruction, as depicted in~\cref{fig:concept}, which aims to reconstruct a \textit{complete geometry} for every timestamp, covering both visible and occluded regions, by leveraging information across the entire temporal sequence. In essence, this task enables temporal aggregation of 3D geometry such that all points observed at different timestamps are aligned and aggregated into a chosen target timestamp.

Formally, given a sequence of $N$ temporally continuous RGB images observing a dynamic 3D scene,
\[
\V I_i \in \mathbb{R}^{3 \times H \times W}, \quad i = 0, \dots, N-1,
\]
and a target (aggregate) timestamp $a \in \{0, \dots, N-1\}$, our goal is to reconstruct the \textit{complete} 3D representation of the scene at time step $a$.

For a given target timestamp $a$, the model aggregates geometric cues from all other frames $\{\V I_i\}_{i \neq a}$, including both static backgrounds and moving objects, to infer a \textit{complete} pointmap $\mathbf{P}^a$. Unlike previous methods that only recover visible geometry, this formulation explicitly reconstructs occluded parts of moving objects that are observed in other frames, thereby implicitly embedding motion causality reasoning across time.

\paragraph{Relation to 3D Tracking.}
Geometric Complete 4D Reconstruction inherently establishes dense temporal consistency across frames by aggregating geometry over time. Through this process, the model implicitly infers the motion of each 3D point. In other words, the temporal alignment used for geometric completion naturally yields consistent point trajectories, enabling 3D tracking as a byproduct of reconstruction. In practice, by iteratively treating each frame $\V I_i$ as the target timestamp $\V I_a$, the model reconstructs temporally coherent geometry at each step, and the accumulated temporal consistency across these reconstructions directly produces continuous 3D motion trajectories for all points.

\subsection{Model Architecture}
\ourwork is built upon a unified transformer architecture that jointly reasons over temporal sequences to reconstruct complete geometry at a chosen target timestamp, while simultaneously estimating per-frame depth and camera parameters. Formally, the model is defined as a mapping:
\begin{equation}
f\left((\V I_i)_{i=0}^{N-1},\, a\right)
=
\left(
\V P_i^a, \bg_i, \, \V D_i 
\right)_{i=0}^{N-1},
\end{equation}
where $\V P_i^a \in \mathbb{R}^{H \times W \times 3}$ represents the aggregated 3D points from frame $i$ toward the target timestamp $a$, $\bg_i \in \mathbb{R}^9$ denotes the camera parameters (including intrinsics and extrinsics)~\cite{wang2024vggsfm}, and $\V D_i \in \mathbb{R}^{H \times W}$ is the predicted depth map. 

Each input image $\V I$ is divided into $K$ patches and embedded into a set of visual tokens $\tok^I \in \mathbb{R}^{K \times C}$ using DINOv2. All subsequent computations are based on these tokens.

\paragraph{Aggregation-aware Token Design.}
To enable geometric aggregation and completion, we introduce a novel design of aggregation tokens $\tok^D$. Specifically, two sets of aggregation tokens are initialized: one set $\tok^D_a$ for the target timestamp and another set $\tok^D_{/a}$ shared among all other timestamps. These tokens are concatenated with each frame’s visual tokens, allowing the model to explicitly identify the aggregation target. By varying the target timestamp $a$, \ourwork learns to gather geometric cues from other frames toward the $a$, thereby reconstructing the complete scene geometry.

Following VGGT, we include camera tokens $\tok^\bg$ and registration tokens $\tok^R$, which encode camera poses and align features to a shared coordinate system, respectively. In particular, we initialize two sets of registration tokens: one $\tok^R_1$ for the first frame and another $\tok^R_{2:N}$ shared by all remaining frames, so that the model learns a unified representation under the coordinate system of the first frame. The aggregation tokens $\tok^D$ further capture frame-specific semantics, facilitating temporal alignment across time. We also experimented with additive fusion instead of concatenation, and detailed comparisons are discussed later.

\paragraph{Attention Mechanism.}
During both the frame-level and global attention stages of the transformer, aggregation tokens from the target frame and the remaining frames interact through self-attention to exchange temporal information. This process progressively aligns the features of all frames toward the target frame, integrating observations from both past and future frames. Consequently, the model reconstructs a holistic 3D representation of the scene and synthesizes complementary viewpoints that are otherwise unobservable from a single frame. 

\paragraph{Prediction Heads.}
We adopt the original camera and depth heads from VGGT and introduce an additional \textit{aggregation head} to enable temporally consistent 4D reconstruction. Both the depth and aggregation heads are implemented using DPT-style decoders. The camera head predicts camera parameters directly from the camera tokens, while the depth head estimates depth maps from patch tokens.

The aggregation head takes the features of each input frame and predicts the corresponding 3D scene representation aligned with the target frame’s timestamp. By combining the outputs of the aggregation head across all frames, we obtain a complete and temporally coherent 3D point map of the scene from the viewpoint of the chosen target frame.

\subsection{Training}

\paragraph{Training Losses.}

We train our \ourwork model $f$ using a multi-task loss:
\begin{equation}
\label{eq:training_loss}
\mathcal{L} =\lambda \mathcal{L}_\text{point}
+ \mathcal{L}_\text{camera}
+ \mathcal{L}_\text{depth}.
\end{equation}

To address the difficulty of supervising dynamic and misaligned regions in 4D completion, 
we introduce a novel \textit{Focal-Weighted Point Loss}. Given predicted points 
$\hat{\mathbf{P}}_i^a$ warped to a target timestamp $a$ and ground truth $\mathbf{P}_i^a$, 
we follow VGGT and use a predicted uncertainty map $\hat{\Sigma}_{i,a}^P$ for 
aleatoric weighting.

To further emphasize hard samples, we adopt a focal-style point weight:
\[
\mathbf{w}_i^a = |\beta \mathbf{e}_i^a|^\gamma,
\qquad 
\mathbf{e}_i^a = \hat{\mathbf{P}}_i^a - \mathbf{P}_i^a,
\]
which increases supervision on points with larger alignment errors. This improves 
robustness in highly dynamic regions.

The final loss for target frame $a$ is:
\begin{equation}
\begin{aligned}
\mathcal{L}_\text{point} &=
\sum_{i=1}^N \Big(
\| \hat{\Sigma}_{i,a}^P \odot \mathbf{w}_i^a 
   \odot (\hat{\mathbf{P}}_i^a - \mathbf{P}_i^a) \| \\
&\quad + \| \hat{\Sigma}_{i,a}^P \odot 
   (\nabla \hat{\mathbf{P}}_i^a - \nabla \mathbf{P}_i^a) \|
   - \alpha \log \hat{\Sigma}_{i,a}^P \Big),
\end{aligned}
\end{equation}
where $\odot$ denotes channel-broadcast multiplication.

This Focal-Weighted Point Loss is a key component of our framework, yielding more accurate 
4D reconstruction in challenging regions.

Following VGGT, we employ the same loss functions to supervise both the camera parameters and depth estimation. 


\paragraph{Training Setup.}

The model is initialized from VGGT, with both the camera and depth heads frozen during training. 
The Aggregation head is introduced in place of the original point head, inheriting its parameters.

Training is performed by minimizing the training loss~\eqref{eq:training_loss} using the AdamW optimizer for $10$ epochs, together with a cosine learning rate schedule featuring a peak learning rate of $1e-5$ and an $0.5$-iteration warmup. 
Input frames, depth maps, and point maps are resized such that their longer side is at most $518$ pixels, with width-to-height ratios randomly sampled between $0.5$ and $3.4$.
We further apply data augmentations including random color jitter, gaussian blur, and grayscale conversion. 
Training is conducted on $8$ A100 GPUs over $23$ hours. 

\paragraph{Training Data.}

We use three datasets for training: Point Odyssey~\cite{zheng2023pointodyssey}, Dynamic Replica~\cite{karaev2023dynamicstereo}, and SAIL-VOS 3D~\cite{hu2021sail}. They consist of dynamic scenes and provide depth maps, camera parameters, and 3D point trajectories. Their rich motion patterns, including people with storylines~\cite{hu2021sail}, make them suitable for learning pixel-level correspondences across space and time. Among them, SAIL-VOS 3D provides the most complete 3D labels, offering meshes at all timestamps, which we use to generate 3D trajectories. We represent each pixel using barycentric coordinates on a mesh tile, and enforce temporal consistency via vertex correspondences across timestamps. In addition, we split the original long sequences into shorter, temporally consistent clips by identifying large depth-shift changes, which correspond to abrupt camera cuts in the storytelling videos. The final processed SAIL-VOS 3D  training dataset contains about 704 unique sequences, combined with 109 sequences in Point Odyssey and 483 sequences in Dynamic Replica, to serve as our whole training dataset.

%% file: sec/4_exp.tex
\section{Experiment}

\begin{table}[t!]
    \centering
    \scriptsize
    \renewcommand{\arraystretch}{1.15}
    \renewcommand{\tabcolsep}{7pt}

    \begin{tabular}{@{}lccccccc@{}}
        \toprule

            & \multicolumn{2}{c}{Acc.$\downarrow$} 
            & \multicolumn{2}{c}{Complet.$\downarrow$} 
            & \multicolumn{2}{c}{N.C.$\uparrow$} \\
        \cmidrule(lr){2-3} \cmidrule(lr){4-5} \cmidrule(lr){6-7}
            \textbf{Methods}  & Mean & Med. & Mean & Med. & Mean & Med. \\
        \midrule
          St4RTrack-seq
            & 0.92 & 0.71 & 3.10 & 0.17 & 0.48 & 0.47 \\
          St4RTrack-pairs
            & 0.94 & 0.77 & 2.67 & 0.14 & 0.46 & 0.44 \\
        \midrule
          \textbf{\ourwork (Ours)}
            & \textbf{0.50} & \textbf{0.37} & \textbf{0.26} & \textbf{0.11} & \textbf{0.49} & \textbf{0.49} \\
        \bottomrule
    \end{tabular}

    \caption{\textbf{4D Complete Reconstruction on SAIL-VOS 3D-test.} 
    We report Accuracy (Acc.), Completion (Complet.), and Normal Consistency (N.C.) on all points; each metric is shown as Mean and Median. Best results are \textbf{bold}.}
    \label{tab:4d_complete_recon}
\end{table}
To assess the performance of \ourwork{}, we conduct experiments on three tasks: 4D Complete Reconstruction (Sec. 4.1), 3D Point Tracking (Sec. 4.2), followed by an ablation study (Sec. 4.3).

\subsection{4D Complete Reconstruction}

We evaluate the 4D complete reconstruction performance of \ourwork{} by measuring Accuracy, Completion, and Normal Consistency on the \ourwork{}-benchmark, to provide a comprehensive assessment.

\paragraph{Datasets.}

To evaluate and compare the 4D complete reconstruction capability of models for the task, we select a subset of the SAIL-VOS 3D validation split as \textit{SAIL-VOS 3D-test} for the benchmark, which contains 44 sequences with the same distribution as the original dataset, with the same ratio of indoor/outdoor conditions and similar spatial–temporal distribution of motion amplitude and patterns. 

\paragraph{Baselines.}

To the best of our knowledge, this task represents a novel problem, and there are no existing methods that can be directly used for comparison. The most similar method is St4RTrack~\cite{feng2025st4rtrack}, which leverages pairwise correspondences for joint tracking and reconstruction. Due to the limitation of pairwise tracking, achieving full 4D reconstruction requires anchoring every frame and subsequently aggregating the tracking results for each timestamp. We adopt this method as our baseline and evaluate the two checkpoints provided for different training modes~\cite{feng2025st4rtrack}.

The 3D point tracking series~\cite{xiao2024spatialtracker, xiao2025spatialtrackerv2} focus on sparse-point tracking and are not designed for reconstruction.
On our task, they cannot perform dense per-pixel tracking on the same hardware and lack supervision for currently occluded regions.
Like St4RTrack, they also require post-aggregation for complete reconstruction. Due to these limitations, which are not fair for these tracking methods in the task, we do not use them as our baselines.

\paragraph{Evaluation Metrics.}

To evaluate 4D complete reconstruction performance, we follow the 3D reconstruction evaluation protocol of CUT3R~\cite{wang2025continuous}. Concretely, we evaluate the 3D reconstruction performance of each selected target frame, which aggregates points from the current and all other observations. For computational efficiency, the points are randomly downsampled.

\paragraph{Results.}

The evaluation results are summarized in~\cref{tab:4d_complete_recon}. Our model consistently surpasses the baselines across all  metrics, establishing a new state-of-the-art on the 4D complete reconstruction task. Specifically, we achieve nearly 50\% relative improvements in Accuracy, and an order-of-magnitude improvement in the mean Completion metric. 
These results highlight the model’s superior performance in recovering complete 3D scenes with consistent geometry.

\subsection{3D Point Tracking}
\label{subsec:3DPointTracking}
\begin{table*}[t]
    \centering
    \resizebox{0.95\textwidth}{!}{
    \begin{tabular}{@{}llcccccccc@{}}
    \toprule
    \multicolumn{2}{c}{} 
        & \multicolumn{2}{c}{PO} 
        & \multicolumn{2}{c}{DR}
        & \multicolumn{2}{c}{ADT}
        & \multicolumn{2}{c}{PStudio} \\
    \cmidrule(lr){3-4} \cmidrule(lr){5-6} \cmidrule(lr){7-8} \cmidrule(lr){9-10}

    \multicolumn{1}{c}{\textbf{Category}} 
        & \multicolumn{1}{c}{\textbf{Methods}}
        & APD$\uparrow$ & EPE$\downarrow$ & APD$\uparrow$ & EPE$\downarrow$ & APD$\uparrow$ & EPE$\downarrow$ & APD$\uparrow$ & EPE$\downarrow$ \\
    \midrule

    \multirow{2}{*}{\textbf{Combinational}} 
      & SpaTracker+RANSAC-Procrustes 
        & 53.77 & 43.58 & 58.58 & 104.44 & 66.49 & 16.00 & 52.05 & 42.66 \\
      & SpaTracker+MonST3R
        & 58.61 & 40.85 & 59.21 & 91.36 & 69.94 & 15.11 & 50.16 & 48.37 \\
    \midrule

    \multirow{3}{*}{\textbf{Feed-forward}}
      & MonST3R 
        & 39.36 & 64.52 & 51.86 & 53.13 & 67.92 & 15.78 & 51.32 & 45.68 \\
      & SpaTracker 
        & 51.20 & 46.95 & 58.65 & 108.28 & 67.65 & 16.28 & 62.59 & 30.94 \\
      & St4RTrack 
        & 68.72 & 29.70 & 68.13 & 29.61 & 75.34 & 12.12 & 69.67 & 26.37 \\
        
    \cmidrule{2-10}

      & \textbf{\ourwork (Ours)} 
        & \textbf{80.17} & \textbf{16.07} & \textbf{80.65} & \textbf{15.99} & \textbf{77.34} & \textbf{9.72} & \textbf{76.88} & \textbf{18.52} \\
    \bottomrule
    \end{tabular}}
    \caption{\textbf{World Coordinate 3D Point Tracking on Dynamic Points.}
    We report both APD and EPE for four datasets: PO (Point Odyssey), DR (Dynamic Replica), ADT (Aria Digital Twin), and PStudio (Panoptic Studio). Best results are \textbf{bold}.}
    \label{tab:3dtracking}
\end{table*}

\begin{figure*}[t!]
    \centering
    \includegraphics[width=0.92\textwidth]{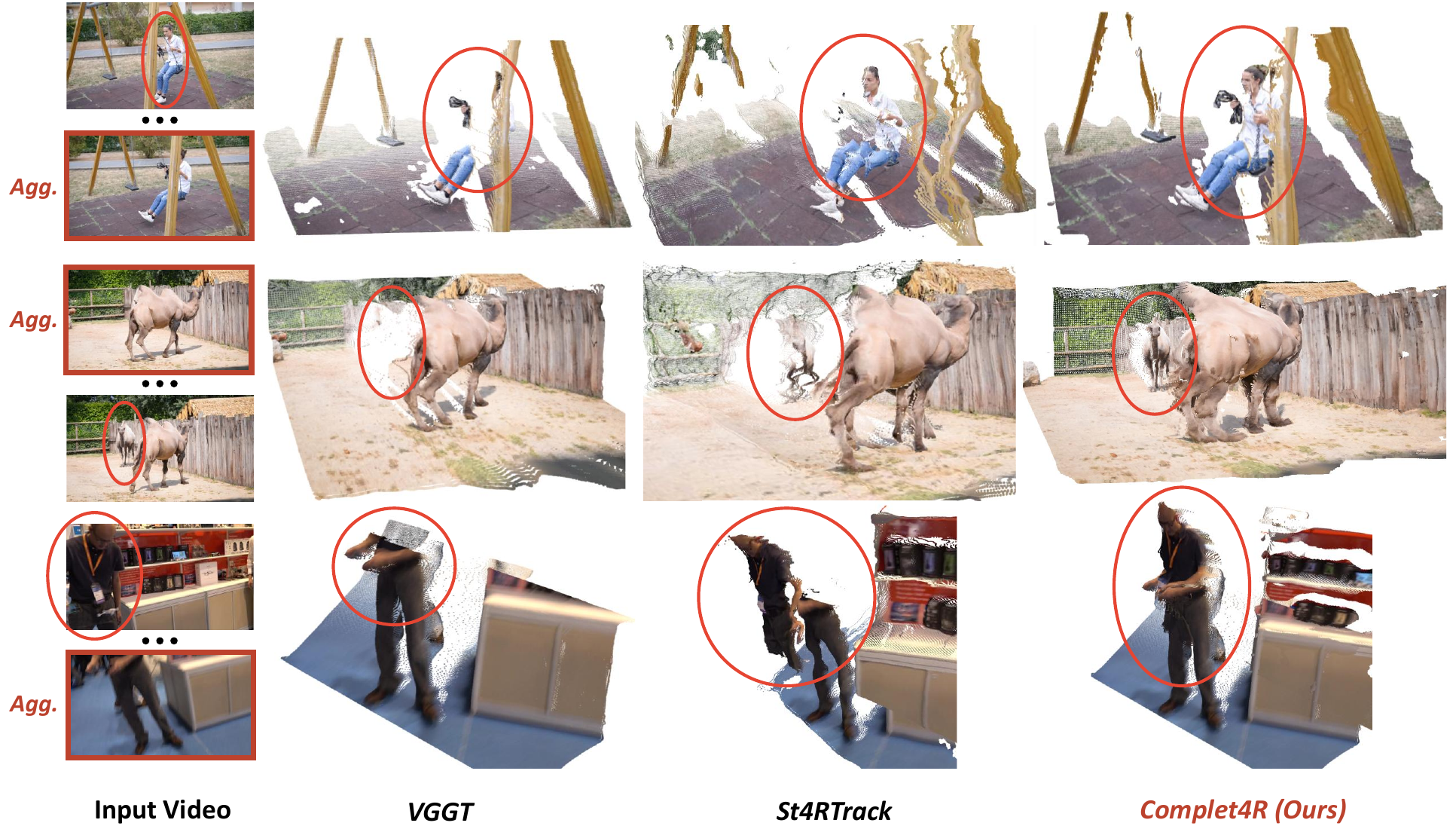}
    \caption{\textbf{Qualitative Results for 4D Complete Reconstruction.} The first column shows the video inputs, with red boxes indicating the target aggregation timestamp for each sequence (\textit{Agg.}: aggregation). 
The subsequent columns present the outputs of different models. 
Our method successfully reconstructs the complete geometry at the target timestamp highlighted by the red ellipses, whereas other methods produce incomplete or geometrically inconsistent reconstructions.}
    \label{fig:vis_recon}
\end{figure*}

\begin{figure*}[t!]
    \centering
    \includegraphics[width=1\textwidth]{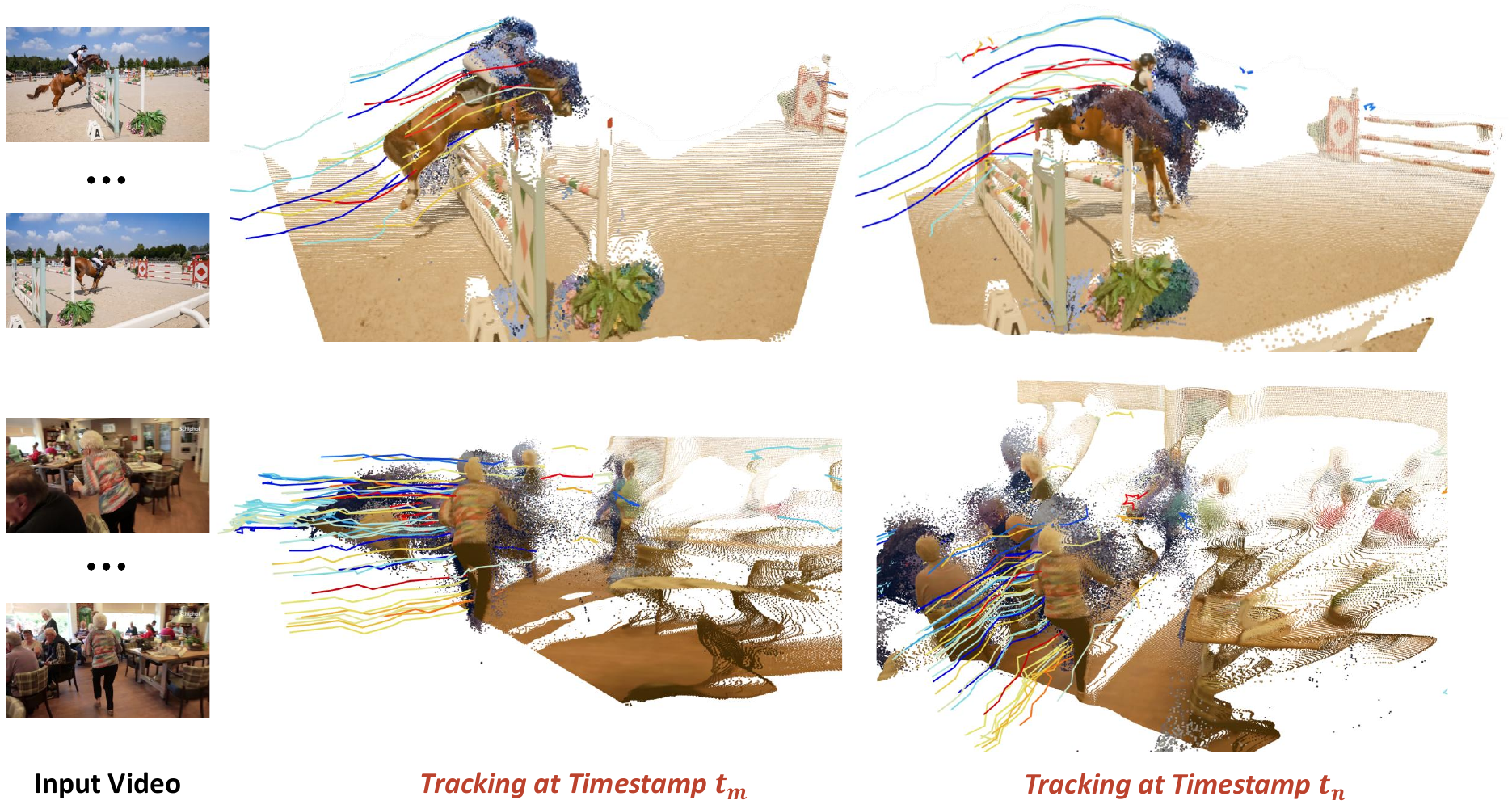}
    \caption{\textbf{Qualitative Results for 3D Dynamic Point Tracking.} The first column shows the input images; the second and third columns display the tracking trajectories produced by our method at successive time steps. The smooth trajectories demonstrate strong spatiotemporal geometric consistency.
    }
    \label{fig:vis_tracking}
\end{figure*}

\begin{table*}[!t]
    \centering
    \normalsize
    \renewcommand{\arraystretch}{1.15}
    \renewcommand{\tabcolsep}{6pt}

    \begin{tabular}{@{}c ccc ccccccc@{}}
        \toprule
        & & & & \multicolumn{2}{c}{Acc.$\downarrow$} 
        & \multicolumn{2}{c}{Complet.$\downarrow$} 
        & \multicolumn{2}{c}{N.C.$\uparrow$} \\
        \cmidrule(lr){5-6} \cmidrule(lr){7-8} \cmidrule(lr){9-10} 
        \textbf{Variants} & \textbf{Loss} & \textbf{Agg. Repre.} & \textbf{Agg. Token} 
        & Mean & Med. & Mean & Med. & Mean & Med. \\
        \midrule
        (1) 
            & Dynamic & - & - & \textbf{0.50} & 0.47 & \textbf{0.26} & 0.20 & 0.48 & 0.47 \\
        (2) 
            & - & Offset & - & 0.63 & 0.42 & 0.50 & 0.11 & 0.44 & 0.40 \\
        (3) 
            & - & - & Add & 0.57 & 0.38 & 0.32 & 0.12 & \textbf{0.50} & \textbf{0.49} \\
        \cmidrule{1-10}
        \textbf{\ourwork (Ours)} & \textbf{Focal} & \textbf{Endpoint} & \textbf{Concatenate} & \textbf{0.50} & \textbf{0.37} & \textbf{0.26} & \textbf{0.11} & 0.49 & \textbf{0.49} \\
        \bottomrule
    \end{tabular}

    \caption{\textbf{Ablation Study on 4D Complete Reconstruction on SAIL-VOS 3D-test.} 
    Accuracy (Acc.), Completion (Complet.), and Normal Consistency (N.C.) are reported over all points; each metric includes Mean and Median. Agg. Repre. means the aggregation representation, Agg. Token means aggregation tokens. Best results are \textbf{bold}. The symbol ``-'' indicates the default setting same as \ourwork.}
    \label{tab:ablation_recon}
\end{table*}

As previously mentioned, \ourwork implicitly supports 3D point tracking. We therefore evaluate its performance on 3D point tracking benchmarks using Average Percent of Points within Delta (APD) and Endpoint Error (EPE), following the protocol of St4RTrack~\cite{feng2025st4rtrack}. Since \ourwork is not specifically designed for 3D point tracking, we perform 4D complete reconstruction for each frame and aggregate corresponding points along the temporal axis to form trajectories for evaluation.

\paragraph{Datasets and Baselines.}

The evaluation is conducted on the \textit{WorldTrack} dataset released by St4RTrack, constructed from Aerial Digital Twin (ADT)~\cite{pan2023aria} and Panoptic Studio~\cite{joo2015panoptic} via TAPVid-3D~\cite{koppula2024tapvid}, and further augmented with PointOdyssey and DynamicReplica. The dataset provides ground-truth 3D trajectories in the world coordinate system with corresponding 2D projections, covering diverse scenarios such as minimal motion, dynamic objects, and large camera motion.

To benchmark the effectiveness of our method, we compare it against the baselines provided by \textit{St4RTrack}~\cite{feng2025st4rtrack}, including \textit{SpaTracker}~\cite{xiao2024spatialtracker}, \textit{MonST3R}~\cite{zhang2024monst3r}, and ~\cite{feng2025st4rtrack}. For fair comparison, we align the camera-coordinate outputs of \textit{SpaTracker} using Procrustes and RANSAC, and transform the predictions of \textit{MonST3R} into world coordinates using its estimated camera poses, enabling 3D tracking evaluation in a unified world coordinate system, which is not natively supported by these methods.

\paragraph{Metric and Results.}

If query points are selected from any frame, \ourwork can predict their 3D coordinates in any other frame. For evaluation consistency, we choose the dynamic visible points in the first frame of each sequence that have complete tracking information as query points.

Next, the predicted 3D points are aligned with the ground truth using the global median. We compute the norms of both predicted points and ground truth points, and derive a scaling factor from their medians for alignment. 

Finally, the aligned predictions are evaluated using APD and EPE. APD is computed as the average percentage of points within four 3D thresholds, $\delta_\text{3D} \in \{0.1\text{m}, 0.3\text{m}, 0.5\text{m}, 1.0\text{m}\}$, across all sequences, with the final score averaged over these thresholds. EPE is calculated as the L2 distance between predicted and ground-truth points, averaged over all frames in the sequences. 

The evaluation results are shown in~\cref{tab:3dtracking}. Although our model was not specifically trained for point tracking, it significantly outperforms the state-of-the-art across all metrics on the point tracking benchmark. 
This demonstrates that our approach can not only reconstruct complete scenes at each timestamp, but also recover temporally consistent and continuous dynamic motions, as further illustrated in~\cref{fig:vis_tracking}.

\subsection{Ablation Study}
\label{subsec:ablation}
We conduct an extensive ablation on the SAIL-VOS 3D-test dataset to analyse three design axes that affect 4D complete reconstruction: the training loss, the aggregation representation, and the processing of aggregation tokens.

\ourwork{} employs a focal-weighted point loss to address the imbalance in point distribution, supervises absolute coordinates at the aggregation timestamp (Endpoint), and fuses aggregation tokens by concatenating them with image patch tokens.

For comparison, we evaluate variants that replace these components as follows; detailed configurations of each variant are provided in the Supplementary Material.
\begin{itemize}
    \item[(1)] Using a dynamic-weighted point loss instead of the focal-weighted point loss, which emphasizes dynamic points by multiplying their loss with a large factor.
    \item[(2)] Supervising displacement relative to a reference frame (Offset) instead of absolute endpoint coordinates.
    \item[(3)] Adding aggregation tokens directly to image patch tokens instead of concatenating them.
\end{itemize}

As shown in Table~\ref{tab:ablation_recon}, our method consistently outperforms these variants across Accuracy, Completion, and Normal Consistency metrics, demonstrating the effectiveness of our design choices.

Specifically, the focal-weighted point loss dynamically adjusts supervision weights via an adaptive mechanism, encouraging the model to focus on challenging regions and thereby improving prediction performance. 
For the aggregation representation, endpoint outperforms offset, and for the aggregation token, concatenation outperforms addition, improving overall reconstruction quality.

%% file: sec/5_conclusion.tex
\section{Conclusion}
\label{conclusion}

In this paper, we introduce \ourwork, a novel end-to-end framework that constructs geometric complete 4D reconstruction for dynamic scenes. 
By jointly reasoning about geometry and motion across all frames, the network recovers geometrically complete and temporally coherent scene reconstruction, including areas occluded in any single view yet visible elsewhere. 
Unlike prior works that stitch pairwise reconstructions or tracks only local motion, \ourwork fuses long-range cues into one unified 4D representation, eliminating drift and ensuring cross-time consistency.
Extensive experiments demonstrate that our method achieves state-of-the-art performance on both the newly proposed 4D complete reconstruction and the 3D point tracking benchmark. We believe that the proposed formulation and method will foster further research, paving the way toward more realistic and consistent world models.

%% file: sec/6_acknowledgment.tex
\section{Acknowledgements}
This work is supported by the National Key R\&D Program of China (2022ZD0161700) and Tsinghua University Initiative Scientific Research Program.

%% file: sec/X_suppl.tex
\clearpage
\setcounter{page}{1}
\maketitlesupplementary


\section{Camera and Depth Supervision}
Our loss function is defined as follows:

\begin{equation}\label{eq:training_loss_supp}
\mathcal{L} = \lambda \mathcal{L}_\text{point}
+ \mathcal{L}_\text{camera}
+ \mathcal{L}_\text{depth}.
\end{equation}
In the following, we provide a detailed description of the camera loss and the depth loss included in the loss function.

The camera loss $\mathcal{L}_\text{camera}$ supervises the predicted camera parameters $\hat{\mathbf{g}}$ by comparing them with the ground-truth parameters $\mathbf{g}$ using the Huber loss $\|\cdot\|_\epsilon$. Specifically, it is defined as:
\begin{equation}
\mathcal{L}_\text{camera} =
\sum_{i=1}^N 
\left\| \hat{\mathbf{g}}_i - \mathbf{g}_i \right\|_\epsilon .
\end{equation}

The depth loss $\mathcal{L}_\text{depth}$ follows VGGT and implements the aleatoric-uncertainty loss, weighting the discrepancy between the predicted depth $\hat{D}_i$ and the ground-truth depth $D_i$ with the predicted uncertainty map $\hat{\Sigma}_i^D$:

\begin{equation}
\begin{aligned}
\mathcal{L}_\text{depth} &= 
\sum_{i=1}^N \Big(
\| \Sigma_i^D \odot (\hat{D}_i - D_i) \| \\
&\quad + \| \Sigma_i^D \odot (\nabla \hat{D}_i - \nabla D_i) \| 
- \alpha \log \Sigma_i^D
\Big)
\end{aligned}
\end{equation}
where $\odot$ is the channel-broadcast element-wise product.

\section{Details of the Ablation Study}

We introduce three variants in our ablation study, focusing on the training loss, the aggregation representation, and the way aggregation tokens are processed.

\paragraph{Dynamic-Weighted Point Loss.} 
First, we detail the design of the \textit{dynamic-weighted point loss}, which aims to emphasize points that exhibit significant motion across timestamps in the sequence.

Given the predicted points $\hat{\mathbf{P}}_i^a$ warped to a target timestamp $a$ and the corresponding ground-truth points $\mathbf{P}_i^a$, the dynamic-weighted point loss function is defined as follows:

\begin{equation}
\begin{aligned}
\mathcal{L}_\text{point} &=
\sum_{i=1}^N \Big(
\| \hat{\Sigma}_{i,a}^P \odot \mathbf{w}_i^a 
   \odot (\hat{\mathbf{P}}_i^a - \mathbf{P}_i^a) \| \\
&\quad + \| \hat{\Sigma}_{i,a}^P \odot 
   (\nabla \hat{\mathbf{P}}_i^a - \nabla \mathbf{P}_i^a) \|
   - \alpha \log \hat{\Sigma}_{i,a}^P \Big),
\end{aligned}
\end{equation}
where $\hat{\Sigma}_{i,a}^P$ is the predicted uncertainty map for aleatoric weighting, $\alpha$ is a regularization coefficient, and the weights \(\mathbf{w}_i^a\) are computed based on whether a point is dynamic:
\begin{equation}
\mathbf{w}_i^a =
\begin{cases}
\text{1000}, & \text{dynamic point}, \\
1, & \text{static point},
\end{cases}
\end{equation}

The method for distinguishing dynamic points from static points is as follows. 
For each point, we first compute its offsets across time, defined as the Euclidean distance between the point's coordinate at any timestamp and its coordinate at the target timestamp $a$:
\begin{equation}
\Delta \mathbf{P}_i^t = \|\mathbf{P}_i^t - \mathbf{P}_i^a\|_2, \qquad t = 0, \dots, N-1,
\end{equation}
where $\mathbf{P}_i^t$ denotes the coordinate of point $i$ at timestamp $t$, and $\mathbf{P}_i^a$ is its coordinate at the target timestamp $a$. A point is considered \textit{dynamic} if its motion exceeds a dataset-specific threshold:

\begin{equation}
\|\Delta \mathbf{P}_i\| > \delta, \qquad \delta = \text{threshold}_{\text{dataset}}.
\end{equation}
Specifically, we set $\delta = 0.03$ for the Point Odyssey and Dynamic Replica datasets, and $\delta = 0.01$ for the SAIL-VOS 3D dataset.

\paragraph{Aggregation Representation.} 
We consider two types of aggregation representations. The first is the predicted 3D coordinates of points at the target timestamp $a$, referred to as the \textit{endpoint}. The second is the predicted offsets from points at each timestamp to their corresponding coordinates at the target timestamp $a$, referred to as the \textit{offset}. 
We describe the supervision method for the \textit{offset} representation below.

The ground-truth offsets are computed as follows:
\begin{equation}
\mathbf{O}_i^t = \mathbf{P}_i^a - \mathbf{P}_i^t, \qquad t = 0, \dots, N-1,
\end{equation}
where $\mathbf{O}_i^t$ represents the ground-truth offset of point $i$ from timestamp $t$ to the target timestamp $a$, 
$\mathbf{P}_i^t$ and $\mathbf{P}_i^a$ denote the 3D coordinates of point $i$ at timestamp $t$ and the target timestamp $a$, respectively.

Given these ground-truth offsets, the point-wise loss is defined to supervise both the positional and smoothness discrepancies, while also incorporating the predicted aleatoric uncertainty $\hat{\Sigma}_{i,a}^O$ and the focal-style point weighting $\mathbf{w}_i^a$: 

\begin{equation}
\begin{aligned}
\mathcal{L}_\text{point} &=
\sum_{i=1}^N \Big(
\| \hat{\Sigma}_{i,a}^O \odot \mathbf{w}_i^a 
   \odot (\hat{\mathbf{O}}_i^a - \mathbf{O}_i^a) \| \\
&\quad + \| \hat{\Sigma}_{i,a}^O \odot 
   (\nabla \hat{\mathbf{O}}_i^a - \nabla \mathbf{O}_i^a) \|
   - \alpha \log \hat{\Sigma}_{i,a}^O \Big).
\end{aligned}
\end{equation}

\begin{table*}[t]
    \centering
    \resizebox{0.95\textwidth}{!}{
    \begin{tabular}{@{}llcccccccc@{}}
    \toprule
    \multicolumn{2}{c}{} 
        & \multicolumn{2}{c}{PO} 
        & \multicolumn{2}{c}{DR}
        & \multicolumn{2}{c}{ADT}
        & \multicolumn{2}{c}{PStudio} \\
    \cmidrule(lr){3-4} \cmidrule(lr){5-6} \cmidrule(lr){7-8} \cmidrule(lr){9-10}

    \multicolumn{1}{c}{\textbf{Category}} 
        & \multicolumn{1}{c}{\textbf{Methods}}
        & APD$\uparrow$ & EPE$\downarrow$ & APD$\uparrow$ & EPE$\downarrow$ & APD$\uparrow$ & EPE$\downarrow$ & APD$\uparrow$ & EPE$\downarrow$ \\
    \midrule

    \multirow{2}{*}{\textbf{Combinational}} 
      & SpaTracker+RANSAC-Procrustes 
        & 61.00 & 33,38 & 61.65 & 37.20 & \textbf{88.65} & 5.96 & 67.82 & 26.60 \\
      & SpaTracker+MonST3R
        & 61.78 & 32.90 & 61.88 & 36.81 & 87.32 & \textbf{4.85} & 64.32 & 29.71 \\
    \midrule

    \multirow{3}{*}{\textbf{Feed-forward}}
      & MonST3R 
        & 48.95 & 47.68 & 55.36 & 38.72 & 84.73 & 7.20 & 64.11 & 30.15 \\
      & SpaTracker 
        & 60.49 & 33.74 & 61.32 & 37.50 & 87.68 & 6.16 & 80.76 & 16.50 \\
      & St4RTrack 
        & 67.43 & 28.70 & 67.90 & 26.27 & 85.34 & 6.88 & 76.97 & 19.69 \\
        
    \cmidrule{2-10}

      & \textbf{\ourwork (Ours)} 
        & \textbf{85.82} & \textbf{12.39} & \textbf{85.30} & \textbf{12.78} & 87.58 & 5.39 & \textbf{85.62} & \textbf{12.49} \\
    \bottomrule
    \end{tabular}}
    \caption{\textbf{World Coordinate 3D Point Tracking on Dynamic Points with Global SIM(3) Alignment.}
    }
    \vspace{-0.5em}
    \label{tab:3dtrackingsim3}
\end{table*}

\paragraph{Aggregation Tokens.}
We design two strategies for handling the aggregation tokens. The first strategy concatenates the aggregation tokens with the image tokens. The second strategy broadcasts the aggregation tokens and adds them to all image tokens.

\section{Additional Experiments}

\subsection{Additional 3D Point Tracking Evaluation}
We additionally test the 3D point tracking on dynamic points in world coordinates with global SIM(3) alignment (SE(3) plus a global scale factor), and the results reported in \cref{tab:3dtrackingsim3}. They are also tested on the four datasets described in the 3D Point Tracking Section. \ourwork{} outperforms all competing approaches by a substantial margin on most datasets, except on ADT, where it achieves performance close to the best method. This demonstrates the strong potential tracking ability of our work.

\subsection{Depth and Camera Pose Estimation}

\begin{table}[t]
\centering
\scriptsize
\resizebox{\linewidth}{!}{
\renewcommand{\arraystretch}{1.0}
\renewcommand{\tabcolsep}{1.4pt}
\begin{tabular}
{@{}cccccccc@{}}
\toprule
& \multicolumn{2}{c}{\textbf{Sintel}} 
& \multicolumn{2}{c}{\textbf{BONN}} 
& \multicolumn{2}{c}{\textbf{KITTI}} \\ 
\cmidrule(lr){2-3} \cmidrule(lr){4-5} \cmidrule(lr){6-7}
\textbf{Method} 
& {Abs Rel $\downarrow$} & {$\delta$\textless $1.25\uparrow$} 
& {Abs Rel $\downarrow$} & {$\delta$\textless $1.25\uparrow$} 
& {Abs Rel $\downarrow$} & {$\delta$\textless $1.25\uparrow$} \\ 
\midrule

VGGT 
& \textbf{0.298} &  \textbf{68.1}
& \textbf{0.057} & \textbf{96.8} 
& \textbf{0.061} & \textbf{97.0} \\

CUT3R
& 0.421 & 47.9
& 0.078 & 93.7
& 0.118 & 88.1 \\

Ours
& \underline{0.353} & \underline{63.9}  
& \underline{0.066} & \underline{95.9} 
& \underline{0.079}  & \underline{94.9}  \\

\bottomrule
\end{tabular}
}
\vspace{-0.8em}
\caption{Evaluation on Video Depth Estimation.}
\vspace{-0.7em}
\label{tab:video_depth}
\end{table}

\begin{table}[t]
\centering
\footnotesize
\renewcommand{\arraystretch}{0.95}
\setlength{\tabcolsep}{1.1pt}
\resizebox{\linewidth}{!}{ 
\begin{tabular}{@{}cccc@{\hspace{10pt}}ccc@{}}
\toprule
& \multicolumn{3}{c}{\textbf{Sintel}} 
& \multicolumn{3}{c}{\textbf{ScanNet}} \\ 
\cmidrule(lr){2-4} \cmidrule(lr){5-7}
\textbf{Method}

& ATE$\downarrow$ & RPE trans $\downarrow$ & RPE rot $\downarrow$
& ATE$\downarrow$ & RPE trans $\downarrow$ & RPE rot $\downarrow$ \\

\midrule
VGGT
& \textbf{0.168} & \textbf{0.065} & \textbf{0.477}
& \textbf{0.035} & \textbf{0.015} & \textbf{0.380} \\

CUT3R
& \underline{0.213} & \underline{0.066} & \underline{0.621} 
& 0.099 & \underline{0.022} &
0.600 \\

Ours
& 0.215 & 0.089 & 1.081
& \underline{0.050} & 0.023 & \underline{0.549} \\

\bottomrule
\end{tabular}
}
\vspace{-0.8em}
\caption{Evaluation on Camera Pose Estimation.}
\vspace{-0.4em}
\label{tab:camera_pose}
\end{table}

We evaluate our method and prior methods on multi-frame depth estimation and camera pose estimation, with results reported in Tables ~\ref{tab:video_depth} and ~\ref{tab:camera_pose}. Although our method does not outperform VGGT, it retains the core capabilities after fine-tuning for 4D complete reconstruction, while achieving competitive or superior performance compared to CUT3R.

\subsection{Runtime and Memory}
\begin{table}[t]
\centering
\scriptsize
\resizebox{\linewidth}{!}{
\setlength{\tabcolsep}{3pt}
\begin{tabular}{c c c c c c c c c}
\toprule
\textbf{Frames} & 2 & 4 & 8 & 16 & 32 & 64 & 128 & 256 \\
\midrule
\textbf{Time (s)} & 0.10 & 0.15 & 0.26 & 0.51 & 1.16 & 2.95 & 8.53 & 27.86 \\
\textbf{Mem. (GB)} & 7.0 & 7.6 & 9.0 & 10.2 & 12.6 & 15.1 & 18.0 & 28.6 \\
\bottomrule
\end{tabular}
}
\vspace{-0.6em}
\caption{Runtime and GPU memory consumption for different sequence lengths. Runtime is measured in seconds,
and GPU memory usage is reported in gigabytes.}
\vspace{-0.6em}
\label{tab:runtime_memory}
\end{table}

As shown in Tab. ~\ref{tab:runtime_memory}, we evaluate the inference runtime and peak GPU memory usage of Complet4R under different numbers of input frames. All experiments are conducted on a single NVIDIA A100 GPU, with an image resolution of 294 × 518.

Despite incorporating additional functionalities and a more complex architecture than VGGT, Complet4R maintains competitive efficiency. For example, inference on 64 frames takes less than 3 seconds with only ~15 GB of peak GPU memory usage, demonstrating the practicality of our model for real-world applications.

Moreover, due to the use of global attention, the runtime does not scale linearly with the number of frames, which poses challenges for long-video inference. A potential solution is to adopt Flash Attention to further optimize the attention layers, which can reduce both computation time and memory consumption.

\section{Additional Qualitative Results}

\begin{figure*}[t!]
    \centering
    \includegraphics[width=\textwidth]{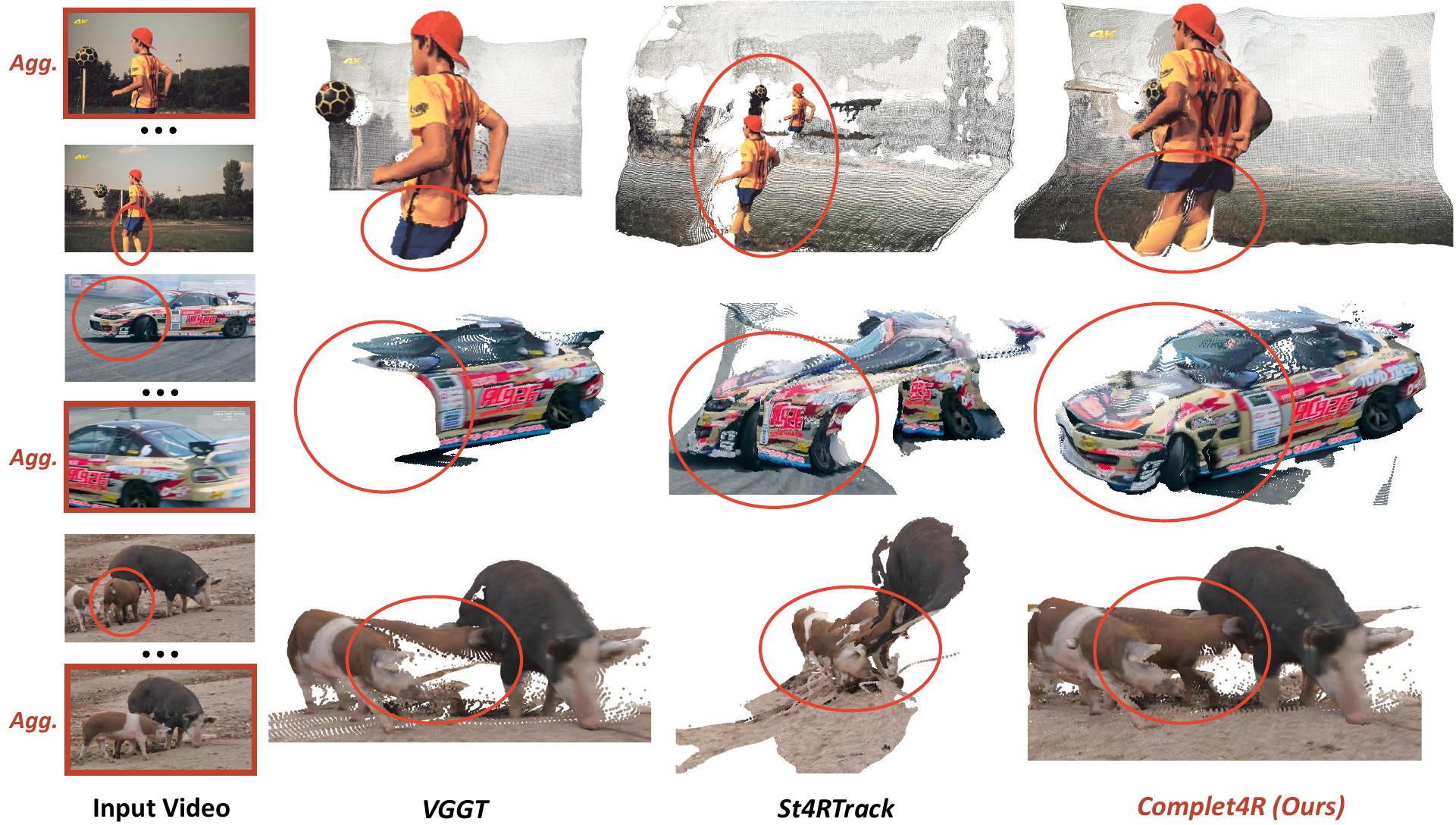}
    \caption{\textbf{More Qualitative Results for 4D Complete Reconstruction.} The first column shows the video inputs, with red boxes indicating the target aggregation timestamp for each sequence (\textit{Agg.}: aggregation). 
The subsequent columns present the outputs of different models. 
Our method successfully reconstructs the complete geometry at the target timestamp highlighted by the red ellipses, whereas other methods produce incomplete or geometrically inconsistent reconstructions.}
    \label{fig:supp-recon}
\end{figure*}

\begin{figure*}[t!]
    \centering
    \includegraphics[width=\textwidth]{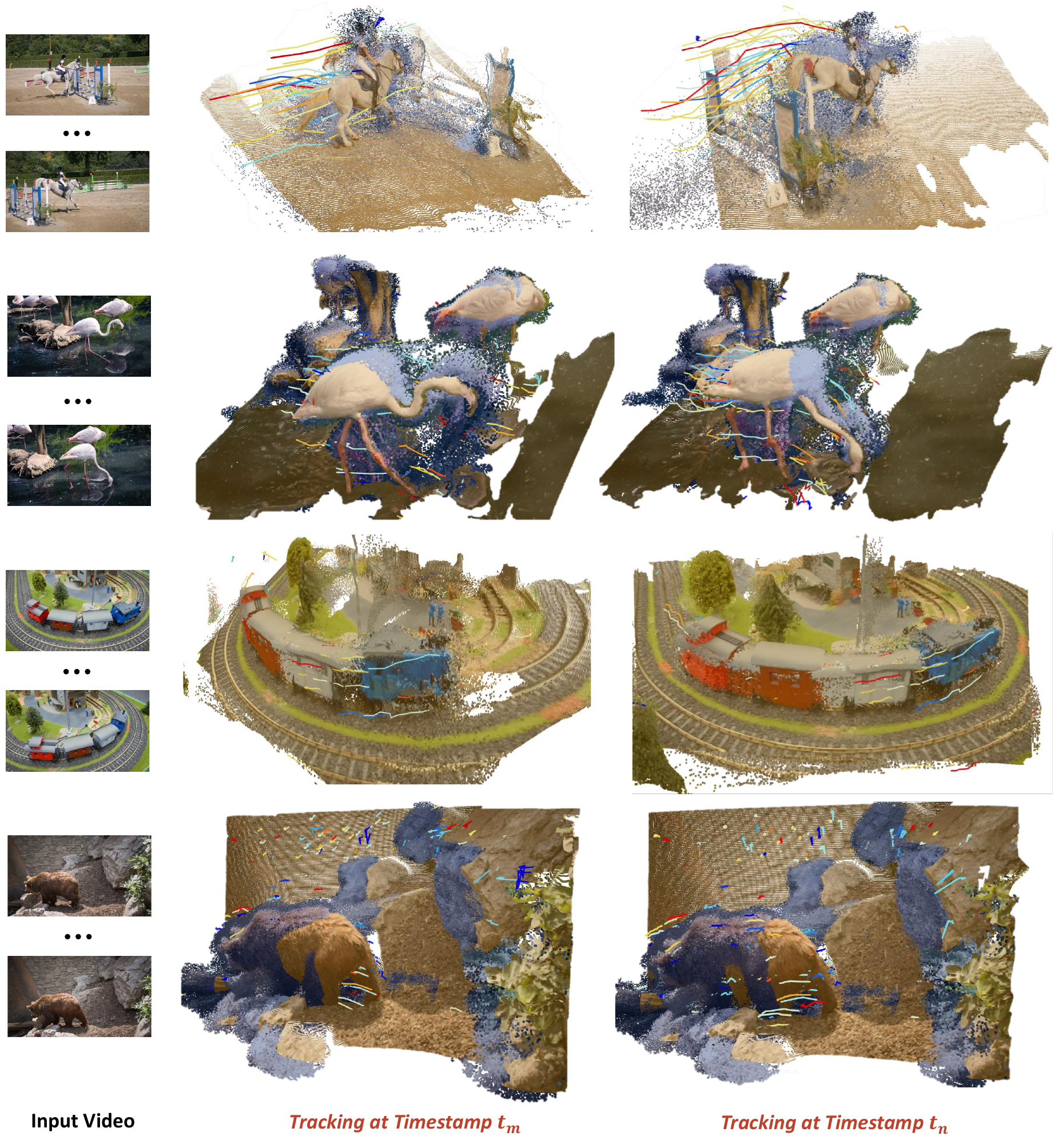}
    \caption{\textbf{More Qualitative Results for 3D Dynamic Point Tracking.} The first column shows the input images; the second and third columns display the tracking trajectories produced by our method at successive time steps. The smooth trajectories demonstrate strong spatiotemporal geometric consistency.
    }
    \label{fig:supp-track}
\end{figure*}

We present additional reconstruction results, as shown in \cref{fig:supp-recon}, and tracking results, as shown in \cref{fig:supp-track}, demonstrating that our model achieves excellent 4D complete reconstruction and 3D point tracking performance.

\section{Additional Discussion}
\begin{figure*}[h]
    \centering
    \includegraphics[width=0.86\textwidth]{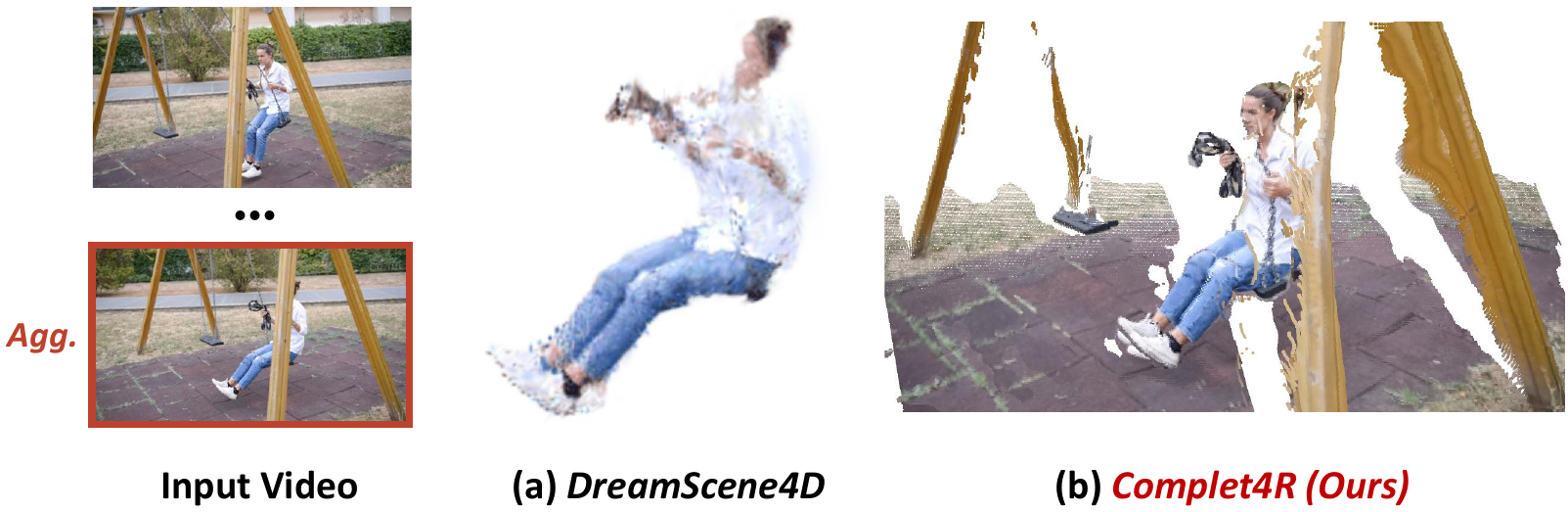}
    \caption{\textbf{Comparison with DreamScene4D.} The left shows the video inputs, while the right presents the reconstruction results: (a) from DreamScene4D, and (b) from our method. Our approach demonstrates higher reconstruction quality.}
    \label{fig:discussion}
\end{figure*}

Existing NeRF/Gaussian-based scene generation methods address related problems. Here, we select the representative Gaussian-based generative approach DreamScene4D for comparison. Such methods aim to obtain visually pleasing renderings through iterative optimization, which differs from our feed-forward regression framework that focuses on geometric consistency.
In complex scenarios, such as the occlusion case shown in ~\ref{fig:discussion}, DreamScene4D exhibits geometric inconsistencies and blurry artifacts, whereas Complet4R demonstrates strong geometric consistency. Moreover, benefiting from its feed-forward design, Complet4R runs significantly faster than DreamScene4D (0.5 s vs. 28 min for 16 frames).

